\documentclass[conference,9pt,a4paper]{IEEEtran}
\IEEEoverridecommandlockouts

\usepackage{cite}
\usepackage{amsmath,amssymb,amsfonts}
\usepackage{algorithmic}
\usepackage{graphicx}

\usepackage{textcomp}
\usepackage{xcolor}

\usepackage{amsthm}
\usepackage{multirow}
\usepackage{array}
\usepackage{color}
\usepackage[utf8]{inputenc}
\usepackage[T1]{fontenc}
\usepackage{float}
\usepackage{booktabs}
\usepackage[font=small,labelfont=bf]{caption}

\ifCLASSOPTIONcompsoc 
 \usepackage[caption=false,font=normalsize,labelfont=sf,textfont=sf]{subfig}
 \else
 \usepackage[caption=false,font=footnotesize]{subfig}
 \fi

\def\0{{\mathbf 0}}
\def\1{{\mathbf 1}}

\def\f{{\mathbf f}}

\def\v{{\mathbf v}}

\def\x{{\mathbf x}}
\def\y{{\mathbf y}}

\def\D{{\mathbf D}}

\def\I{{\mathbf I}}

\def\L{{\mathbf L}}
\def\M{{\mathbf M}}

\def\V{{\mathbf V}}
\def\W{{\mathbf W}}

\def\ie{{\textit{i.e.}}}
\def\eg{{\textit{e.g.}}}

\def\cE{{\mathcal E}}

\def\cG{{\mathcal G}}

\def\cN{{\mathcal N}}
\def\cO{{\mathcal O}}

\def\cR{{\mathcal R}}

\def\cX{{\mathcal X}}

\def\balpha{{\boldsymbol \alpha}}

\def\0{{\mathbf 0}}
\def\1{{\mathbf 1}}

\def\f{{\mathbf f}}

\def\v{{\mathbf v}}

\def\x{{\mathbf x}}
\def\y{{\mathbf y}}

\def\D{{\mathbf D}}

\def\I{{\mathbf I}}

\def\L{{\mathbf L}}
\def\M{{\mathbf M}}

\def\V{{\mathbf V}}
\def\W{{\mathbf W}}

\def\ie{{\textit{i.e.}}}
\def\eg{{\textit{e.g.}}}

\def\cE{{\mathcal E}}

\def\cG{{\mathcal G}}

\def\cN{{\mathcal N}}
\def\cO{{\mathcal O}}

\def\cR{{\mathcal R}}

\def\cX{{\mathcal X}}

\def\cO{{\mathcal O}}
\def\cR{{\mathcal R}}

\def\cN{{\mathcal N}}

\def\balpha{{\boldsymbol \alpha}}

\begin{document}

\title{Unsupervised Graph Spectral Feature Denoising for Crop Yield Prediction\\
\thanks{Gene Cheung acknowledges the support of the NSERC grants RGPIN-2019-06271,  RGPAS-2019-00110.}
}

\author{\IEEEauthorblockN{Saghar Bagheri}
\IEEEauthorblockA{\textit{Dept. of EECS} \\
\textit{York University}\\
Toronto, Canada\\
sagharb@yorku.ca}
\and
\IEEEauthorblockN{Chinthaka Dinesh}
\IEEEauthorblockA{\textit{Dept. of EECS} \\
\textit{York University}\\
Toronto, Canada \\
dineshc@yorku.ca}
\and

\IEEEauthorblockN{Gene Cheung}
\IEEEauthorblockA{\textit{Dept. of EECS} \\
\textit{York University}\\
Toronto, Canada \\
genec@yorku.ca}
\and
\IEEEauthorblockN{Timothy Eadie}
\IEEEauthorblockA{\textit{} \\
\textit{GrowersEdge}\\
Iowa, USA\\
Timothy.Eadie@growersedge.com}
}
%

\maketitle
%
%
%

%
\begin{abstract}
Prediction of annual crop yields at a county granularity is important for national food production  and price stability.
In this paper, towards the goal of better crop yield prediction, leveraging recent graph signal processing (GSP) tools to exploit spatial correlation among neighboring counties, we denoise relevant features via graph spectral filtering that are inputs to a deep learning prediction model.
Specifically, we first construct a combinatorial graph with edge weights that encode county-to-county similarities in soil and location features via metric learning.
We then denoise features via a maximum a posteriori (MAP) formulation with a graph Laplacian regularizer (GLR).
We focus on the challenge to estimate the crucial weight parameter $\mu$, trading off the fidelity term and GLR, that is a function of noise variance in an unsupervised manner.
We first estimate noise variance directly from noise-corrupted graph signals using a graph clique detection (GCD) procedure that discovers locally constant regions.
We then compute an optimal $\mu$ minimizing an approximate mean square error function via bias-variance analysis.
Experimental results from collected USDA data show that using denoised features as input, performance of a crop yield prediction model can be improved noticeably. 
\end{abstract}
\section{Introduction}
\label{sec:intro}
As weather patterns become more volatile due to unprecedented climate change, accurate \textit{crop yield prediction}---forecast of agriculture production such as corn or soybean at a county / state granularity---is increasingly important in agronomics to ensure a robust and reliable national food supply \cite{cai17}.
A conventional crop yield prediction scheme gathers \textit{relevant features} that influence crop production---\eg, soil composition, precipitation, temperature---as input to a deep learning (DL) model such as convolutional neural net (CNN) \cite{khaki20} and long short-term memory (LSTM) \cite{sun19} to estimate yield per county / state in a \textit{supervised} manner. 
While this is feasible when the training dataset is sufficiently large, the trained model is nonetheless susceptible to noise in feature data, typically collected by USDA from satellite images and farmer surveys\footnote{https://www.usda.gov/}.
In this paper, we focus on the problem of pre-denoising relevant features prior to DL model training to improve crop yield prediction performance. 

Given that basic environmental conditions such as soil makeup, rainfall and drought index at one county are typically similar to nearby ones, one would expect crucial features directly related to crop yields, such as \textit{normalized difference vegetation index} (NDVI) and \textit{enhanced vegetation index} (EVI)~\cite{matsushita2007}, at neighboring counties to be similar as well.
To exploit these inter-county similarities for feature denoising, leveraging recent rapid progress in \textit{graph signal processing} (GSP) \cite{ortega18ieee,cheung18} we pursue a graph spectral filtering approach.
While graph signal denoising is now well studied in many contexts, including general band-limited graph signals \cite{chen15}, 2D images \cite{pang17,vu21}, and 3D point clouds \cite{zeng20,dinesh20}, our problem setting for crop feature denoising is particularly challenging because of its \textit{unsupervised} nature.
Specifically, an obtained feature $\y \in \mathbb{R}^N$ for $N$ counties is typically noise-corrupted, and one has no access to ground truth data $\x^o$ nor knowledge of the noise variance $\sigma^2$. 
Thus, the important weight parameter $\mu$ that trades off the fidelity term $\|\y - \x\|^2_2$ against the graph signal prior such as the \textit{graph Laplacian regularizer}\footnote{$\L$ is the combinatorial graph Laplacian matrix; definitions are formally defined in Section\;\ref{sec:preli}.} $\x^\top \L \x$  \cite{pang17} or \textit{graph total variation} (GTV) \cite{bai19} in a \textit{maximum a posteriori} (MAP) formulation cannot be easily derived \cite{chen17} or trained end-to-end \cite{vu21} as previously done.

\begin{figure}[t]
\centering
\includegraphics[width=0.65\columnwidth]{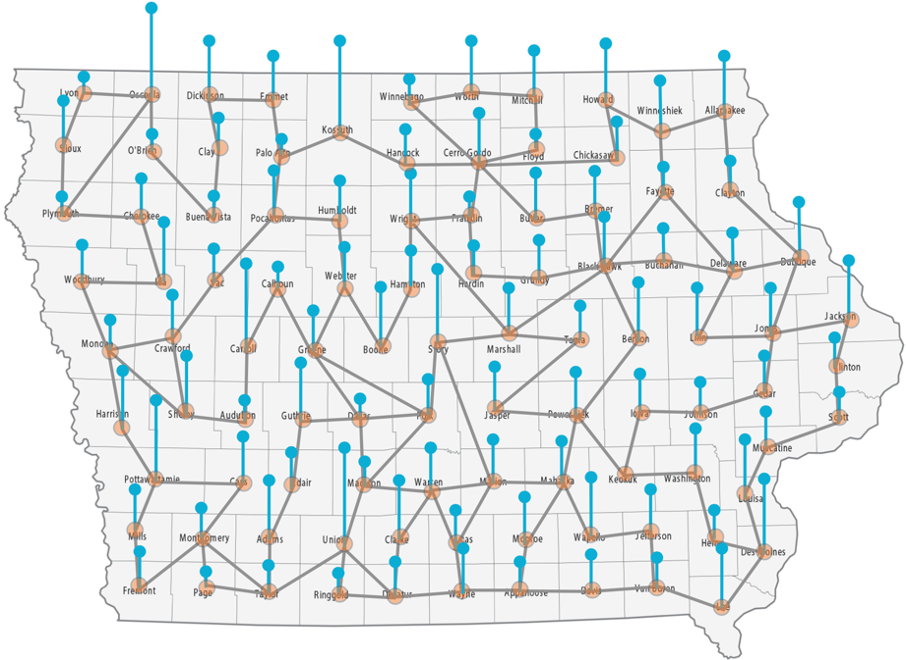}
\vspace{-0.05in}
\caption{Feature of different counties in Iowa as a discrete signal on a combinatorial graph.
}
\label{fig:countygraphsignal}
\end{figure}

In this paper, we focus on the unsupervised estimation of the weight parameter $\mu$ in a GLR-regularized MAP formulation for relevant feature pre-denoising to improve crop yield prediction. 
Specifically, we first construct a combinatorial graph $\cG$ with edge weights $w_{i,j}$ encoding similarities between counties (nodes) $i$ and $j$. 
$w_{i,j}$ is inversely proportional to the \textit{Mahanalobis distance}
$d_{i,j} = (\f_i - \f_j)^\top \M (\f_i - \f_j)$, where $\f_i$ is a vector for node $i$ composed of soil and location features, and $\M$ is an optimized metric matrix \cite{wei_TSP2020}.
We then estimate noise variance $\sigma^2$ directly from noise-corrupted features using our proposed \textit{graph clique detection} (GCD) procedure, generalized from noise estimation in 2D imaging~\cite{wu2015}.
Finally, we derive equations analyzing the \textit{bias-variance tradeoff} \cite{chen17} to minimize the resulting MSE of our MAP estimate and compute the optimal weight parameter $\mu$.
See Fig.\;\ref{fig:countygraphsignal} for an illustration of a similarity graph connecting neighboring counties in Iowa with undirected edges, where the set of feature values per county is shown as a discrete signal on top of the graph. 

Using USDA corn data from $10$ states in the corn belt (Iowa, Illinois, Indiana, Ohio, Nebraska, Minnesota, Wisconsin, Michigan, Missouri, and Kentucky) containing $938$ counties, experimental results show that using our GLR-regularized denoiser with optimized $\mu$ to denoise two important EVI features led to improved performance in an DL model \cite{chen16}: reduction of \textit{root mean square error} (RMSE) \cite{pasquel22} by $0.434\%$ in crop yield prediction compared to the baseline when the features were not pre-denoised. 

The paper is organized as follows.
We first overview our crop yield prediction model in Section\;\ref{sec:overview}.
We then present our unsupervised feature denoising algorithm in Section\;\ref{sec:opt}.
Finally, we present experimental results and conclusion in Section\;\ref{sec:results} and \ref{sec:conclude}, respectively.

\section{Prediction Framework}
\label{sec:overview}
We overview a conventional crop yield prediction framework \cite{khaki20,sun19}.
First, \textit{relevant features} at a county level such as silt / clay / sand percentage in soil, accumulated rainfall, drought index, and growing degree days (GDD) are collected from various sources, including USDA and satellite images.
Features of the same county are inputted to a DL model like CNN or LSTM for future yield prediction, trained in a supervised manner using annual county-level yield data provided by USDA. 
Note that existing yield prediction schemes \cite{khaki20,sun19} focus mainly on exploiting \textit{temporal correlation} (both short-term and long-term) to predict future crop yields.

Depending on the type of features, the acquired measurements may be noise-corrupted. 
This may be due to measurement errors by faulty mechanical instruments, human errors during farmer surveys, etc. 
Given that environmental variables are likely similar in neighboring counties, one would expect similar basic features in a local region. 
To exploit this \textit{spatial correlation} for feature denoising, we employ a graph spectral approach to be described next. 

\section{Feature Denoising}
\label{sec:opt}
\subsection{Preliminaries}
\label{sec:preli}

An $N$-node undirected positive graph $\cG(\cN,\cE,\W)$ can be specified by a symmetric \textit{adjacency matrix} $\W \in \mathbb{R}^{N \times N}$, where $W_{i,j} = w_{i,j} > 0$ is the weight of an edge $(i,j) \in \cE$ connecting nodes $i, j \in \cN = \{1, \ldots, N\}$, and $W_{i,j} = 0$ if there is no edge $(i,j) \not\in \cE$. 
Here we assume there are no self-loops, and thus $W_{i,i} = 0, \forall i$. 
Diagonal \textit{degree matrix} $\D \in \mathbb{R}^{N \times N}$ has diagonal entries $D_{i,i} = \sum_j W_{i,j}$. 
We can now define the combinatorial \textit{graph Laplacian matrix} $\L \triangleq \D - \W$, which is \textit{positive semi-definite} (PSD) for a positive graph $\cG$ (\ie, $W_{i,j} \geq 0, \forall i,j$) \cite{cheung18}.

An assignment of a scalar $x_i$ to each graph node $i \in \cN$ composes a \textit{graph signal} $\x \in \mathbb{R}^N$.
Signal $\x$ is smooth with respect to (w.r.t.) graph $\cG$ if its variation over $\cG$ is small. 
A popular graph smoothness measure is the \textit{graph Laplacian regularizer} (GLR) $\x^\top \L \x$ \cite{pang17}, \ie, $\x$ is smooth iff $\x^\top \L \x$ is small.
Denote by $(\lambda_i,\v_i)$ the $i$-th eigen-pair of matrix $\L$, and $\V$ the eigen-matrix composed of eigenvectors $\{\v_i\}_{i=1}^N$ as columns.
$\V^\top$ is known as the \textit{Graph Fourier Transform} (GFT) \cite{cheung18} that converts a graph signal $\x$ to its graph frequency representation $\balpha = \V^\top \x$. 
GLR can be expanded as
\begin{align}
\x^\top \L \x = \sum_{(i,j) \in \cE} w_{i,j} (x_i - x_j)^2 = \sum_{k=1}^{N} \lambda_k \alpha_k^2 .
\end{align}
Thus, a small GLR means that a connected node pair $(i,j) \in \cE$ with large edge weight $w_{i,j}$ has similar sample values $x_i$ and $x_j$ in the nodal domain, and most signal energy resides in low graph frequency coefficients $\alpha_k$ in the spectral domain---$\x$ is a \textit{low-pass} (LP) signal.

\subsection{Graph Metric Learning}
\label{sec:graph_learning}

Assuming that each node $i \in \cN$ is endowed with a length-$K$ \textit{feature vector} $\f_i \in \mathbb{R}^K$, one can compute edge weight $w_{i,j}$ connecting nodes $i$ and $j$ in $\cG$ as
\begin{align}
w_{i,j} = \exp \left\{ - (\f_i - \f_j)^\top \M (\f_i - \f_j) \right\}    
\end{align}
where $\M \succeq 0$ is a PSD \textit{metric matrix} that determines the square Mahalanobis distance (feature distance) 
$d_{i,j} = (\f_i - \f_j)^\top \M (\f_i - \f_j) \geq 0$ between nodes $i$ and $j$. 
There exist \textit{metric learning} schemes \cite{wei_TSP2020,yang21} that optimize $\M$ given an objective function $f(\M)$ and training data $\cX = \{\x_1, \ldots, \x_T\}$. 
For example, we can define $f(\M)$ using GLR and seek $\M$ by minimizing $f(\M)$:
\begin{align}
\min_{\M \succeq 0} f(\M) = \sum_{t=1}^T \x_t^\top \L(\M) \x_t .
\end{align}
In this paper, we adopt an existing metric learning scheme \cite{wei_TSP2020}, and use soil- and location-related features---clay percentage, available water storage estimate (AWS), soil organic carbon stock estimate (SOC) and 2D location features---to compose $\f_i \in \mathbb{R}^5$.
These features are comparatively noise-free and thus reliable.
We use also these features as training data $\cX$ to optimize $\M$, resulting in graph $\cG$. 
(Node pair $(i,j)$ with distance $d_{i,j}$ larger than a threshold has no edge $(i,j) \not\in \cE$). 
We will use $\cG$ to denoise two EVI features that are important for yield prediction.

\subsection{Denoising Formulation}

Given a constructed graph $\cG$ specified by a graph Laplacian matrix $\L$, one can denoise a target input feature $\y \in \mathbb{R}^N$ using a MAP formulation regularized by GLR \cite{pang17}:
\begin{align}
\min_{\x} \|\y - \x \|^2_2 + \mu \x^\top \L \x
\label{eq:MAP}
\end{align}
where $\mu > 0$ is a weight parameter trading off the fidelity term and GLR. 
Given $\L$ is PSD, objective \eqref{eq:MAP} is convex with a system of linear equations as solution:
\begin{align}
\left(\I + \mu \L \right) \x^* = \y .
\label{eq:MAP_sol}
\end{align}
Given that matrix $\I + \mu \L$ is symmetric, \textit{positive definite} (PD) and sparse, \eqref{eq:MAP_sol} can be solved using \textit{conjugate gradient} (CG)\cite{shewchuk1994} without matrix inverse. 
We focus on the selection of $\mu$ in \eqref{eq:MAP} next.

\subsection{Estimating Noise Variance}

In our feature denoising scenario, we first estimate the noise variance $\sigma^2$ directly from noisy feature (signal) $\y$, using which weight parameter $\mu$ in \eqref{eq:MAP} is computed.
We propose a noise estimation procedure called \textit{graph clique detection} (GCD) when a graph $\cG$ encoded with inter-node similarities is provided. 

We generalize from a noise estimation scheme for 2D images~\cite{wu2015}. 
First, we identify \textit{locally constant regions} (LCRs) $\cR_m$ where signal samples are expected to be similar, \ie, $x_i \approx x_j, \forall i,j \in \cR_m$. 
Then, we compute mean $\bar{x}_m = \frac{1}{|\cR_{m}|}\sum_{i \in \cR_m} x_i$ and variance $\sigma^2_m = \frac{1}{|\cR_{m}|} \sum_{i \in \cR_m} (x_i - \bar{x}_m)^2$ for each $\cR_m$.
Finally, we compute the global noise variance as the weighted average:
\begin{align}
\sigma^2 = \sum_m \frac{|\cR_m|}{\sum_k |\cR_k|}\sigma_m^2 .
\label{eq:noise_variance}
\end{align}

The crux thus resides in the identification of LCRs in a graph. 
 Note that this does not imply conventional \textit{graph clustering}~\cite{schaeffer2007graph}: grouping of \textit{all} graph nodes to two or more non-overlapping sets. 
There is no requirement here to put every node in a LCR.

We describe our proposal based on cliques. 
A \textit{clique} is a (sub-)graph where every node is connected with every other node in the (sub-)graph. 
Thus, a clique implies a node cluster with strong inter-node similarities, which we assume is roughly constant. 
Given an input graph $\cG$, we identify cliques in $\cG$ as follows.

\subsubsection{$k$-hop Connected Graph}

\begin{figure}[t]
\centering
\subfloat[]{
\includegraphics[width=0.21\textwidth]{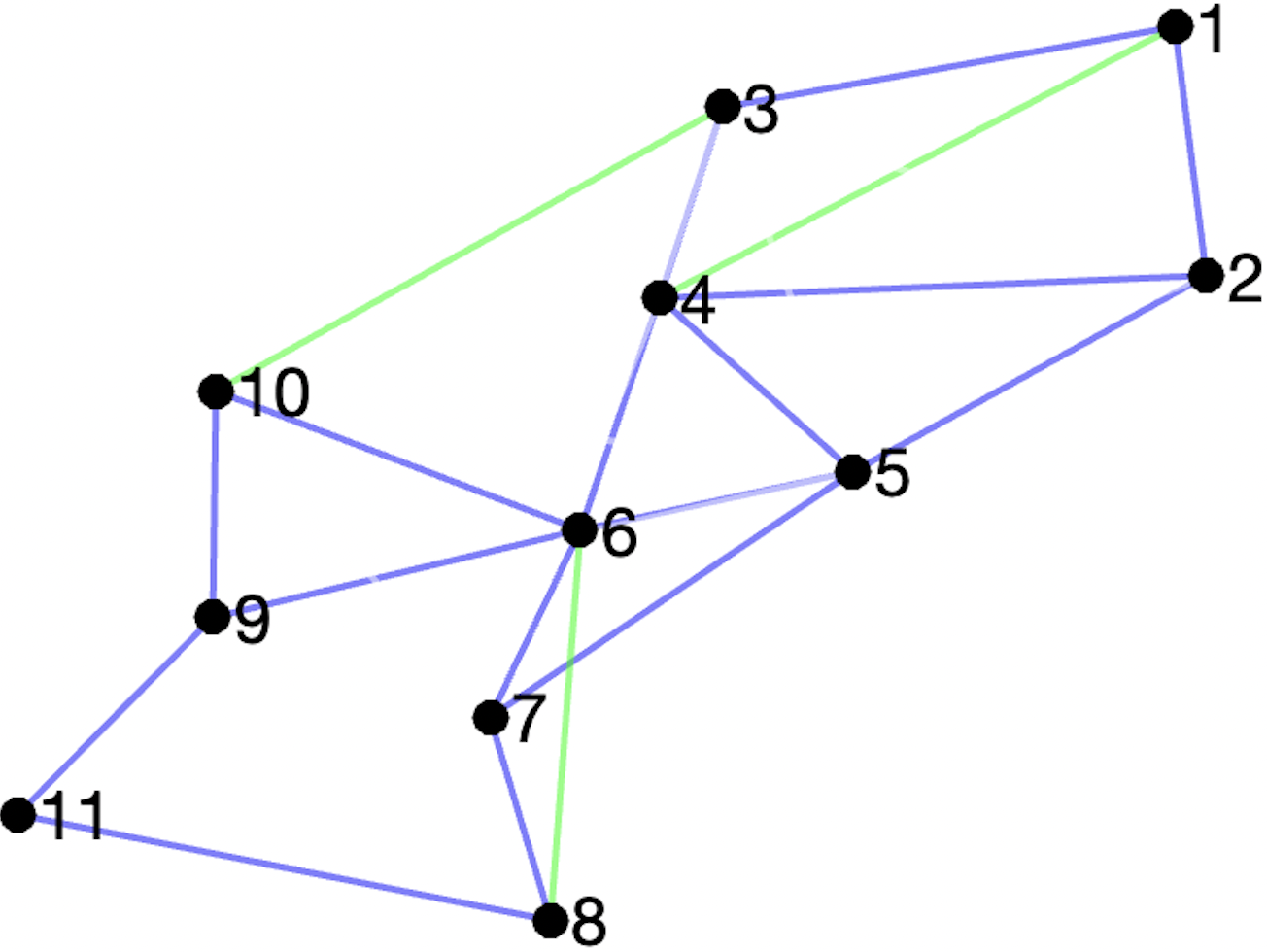}
\label{fig:pcd-on}}
\hspace{-10pt}
\subfloat[{}]{
\includegraphics[width=0.24\textwidth]{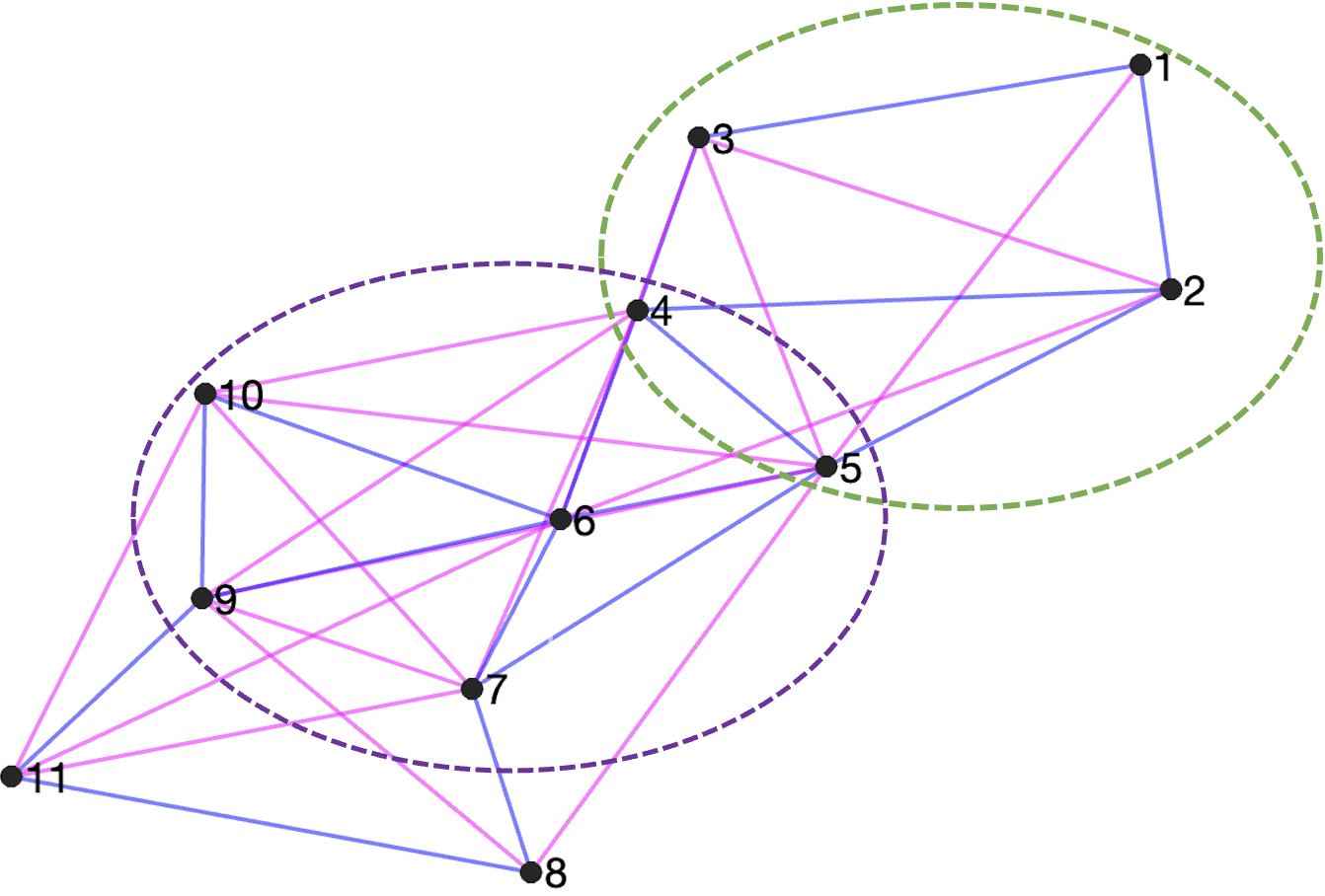}
\label{fig:pcd-off}}
\vspace{-0.05in}
\caption{(a) Example of a $10$-node graph, where edges with weights less than threshold $\hat{w}$ are colored in green; (b) The resulting k-hop connected graph (KCG) for $k=2$, after removing green edges and creating edges (colored in magenta) by connecting 2-hop neighbors. 
Two maximal cliques (out of 5) in KCG are highlighted. } 
\label{fig:cliques}
\end{figure}

We first sort $M = |\cE|$ edges in $\cG$ in weights from smallest to largest. 
For a given \textit{threshold weight} $\hat{w}$ (to be discussed) and $k \in \mathbb{Z}^+$, we remove all edges $(i,j) \in \cE$ with weights $w_{i,j} < \hat{w}$ and construct a \textit{$k$-hop connected graph} (KCG)  $\cG^{(k)}$ with edges connecting nodes $i$ and $j$ that are $k$-hop neighbors in $\cG$.
If $\cG^{(k)}$ has at least a target $\hat{E}$ number of edges, then it is a \textit{feasible} KCG, with minimum connectivity 
$C(\cG^{(k)}) = \hat{w}^k$.
$C(\cG^{(k)})$ is the weakest possible connection between two connected nodes in $\cG^{(k)}$, interpreting edge weights $w_{i,j}$ as conditional probabilities as done in \textit{Gaussian Markov Random Field} (GMRF)~\cite{rue2005}.  

To find threshold $\hat{w}$ for a given $k$, we seek the \textit{largest} $\hat{w}$ for feasible graphs $\cG^{(k)}$ (with minimum $\hat{E}$ edges) via binary search among $M$ edges in complexity $\cO(\log M)$. 
We initialize $k=1$, compute threshold $\hat{w}$, then increment $k$ and repeat the procedure until we identify a maximal\footnote{Given $0<\hat{w}<1$, $\hat{w}^k$ becomes smaller as $k$ increases. Thus, in practice we observe a local maximum in $C(\cG^{(k)})$ as function of $k$.} $C(\cG^{(k)}) = \hat{w}^k$ for $k \in \{1, 2. \ldots\}$.

See Fig.\;\ref{fig:cliques}(b) for an example of a KCG $\cG^{(2)}$ given original graph $\cG$ in Fig.\;\ref{fig:cliques}(a). 
We see, for example, that edge $(3,10)$ is removed from $\cG$, but edge $(4,10)$ is added in $\cG^{(2)}$ because nodes $4$ and $10$ are $2$-hop neighbors in $\cG$. 
The idea is to identify strongly similar pairs in original $\cG$ and connect them with explicit edges in $\cG^{(k)}$. 
Then the maximal cliques\footnote{A maximal clique is a clique that cannot be extended by including one more adjacent node. Hence, a maximal clique is not a sub-set of a larger clique in the graph.} are discovered using algorithm in \cite{cazals2008note}, as shown in Fig.\;\ref{fig:cliques}(b).
The cliques in the resulting graph $\cG^{(k)}$ are LCRs used to calculate the noise variance via \eqref{eq:noise_variance}.

\subsubsection{Target $\hat{E}$ Edges}

The last issue is the designation of targeted $\hat{E}$ edges.
$\hat{E}$ should be chosen so that each clique $m$ discovered in KCG $\cG^{(k)}$ has enough nodes to reliably compute mean $\bar{x}_m$ and variance $\sigma_m^2$. 
We estimate $\hat{E}$ as follows.
Given input graph $\cG$, we first compute the average degree $\bar{d}$. 
Then, we target a given clique to have an average of $n_c$ nodes---large enough to reliably compute mean and variance. 
Thus, the average degree of the resulting graph can be approximated as $\bar{d}+n_c-1$. 
Finally, we compute $\hat{E} \approx N(\bar{d}+n_c-1)$, where $N$ is the number of nodes in the input graph $\cG$.

\subsection{Deriving Weight Parameter} 

\begin{figure}[t]
\centering
\includegraphics[width=0.5\columnwidth]{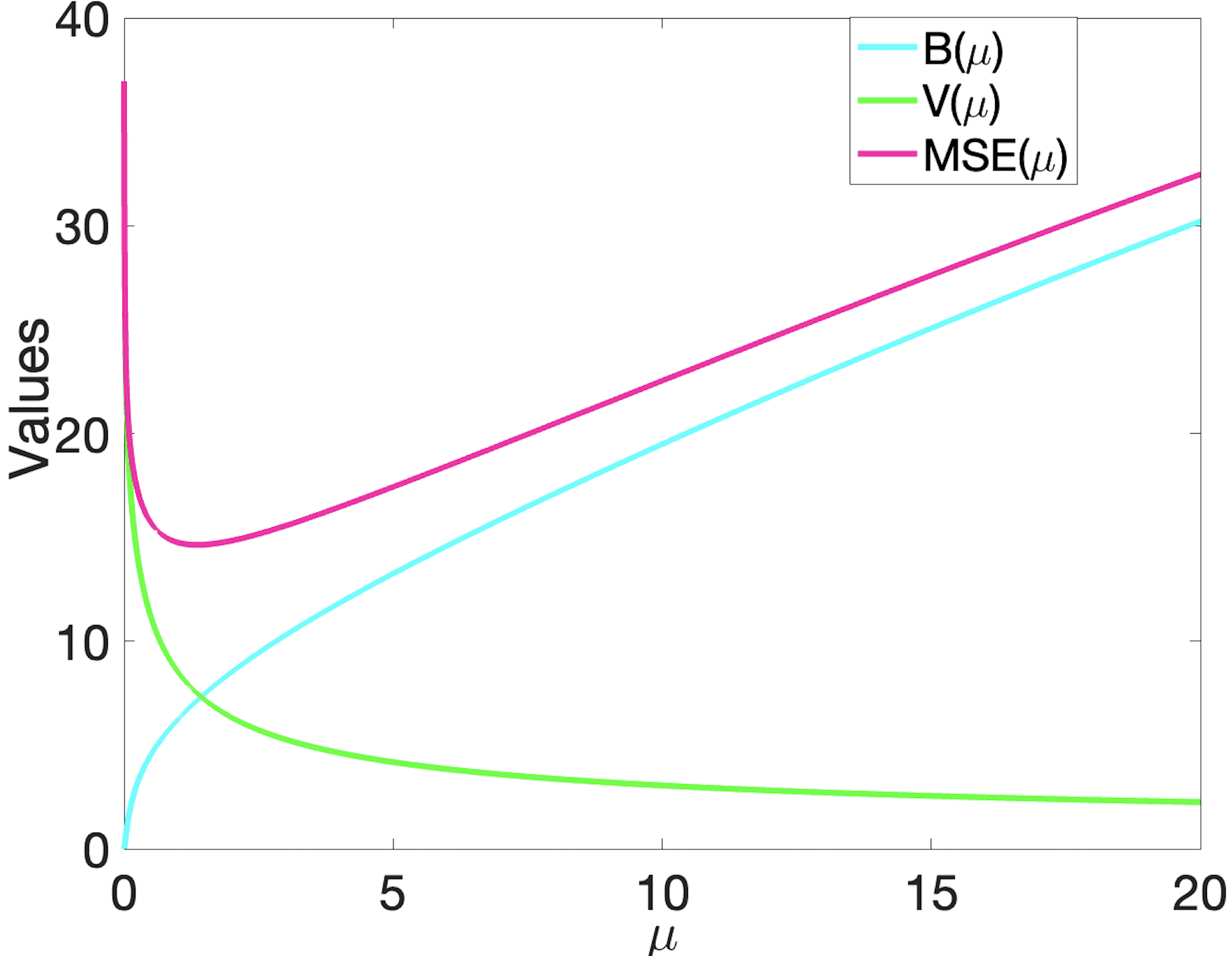}
\caption{Bias $B(\mu)$, variance $V(\mu)$, and MSE($\mu$), as functions of weight parameter $\mu$, for a signal with respect to a graph constructed in Section~\ref{sec:graph_learning}. The underlying graph is constructed by connecting adjacent counties in Iowa. 
Graph signal $\x^{o}$ is the clay percentages in each county. 
We assume $\sigma^2=1$ when computing $B(\mu)$, $V(\mu)$, and MSE($\mu$) in \eqref{eq:MSE}.
}
\label{fig:functions}
\end{figure} 

\begin{figure}[t]
\centering
\subfloat[]{
\includegraphics[width=0.235\textwidth]{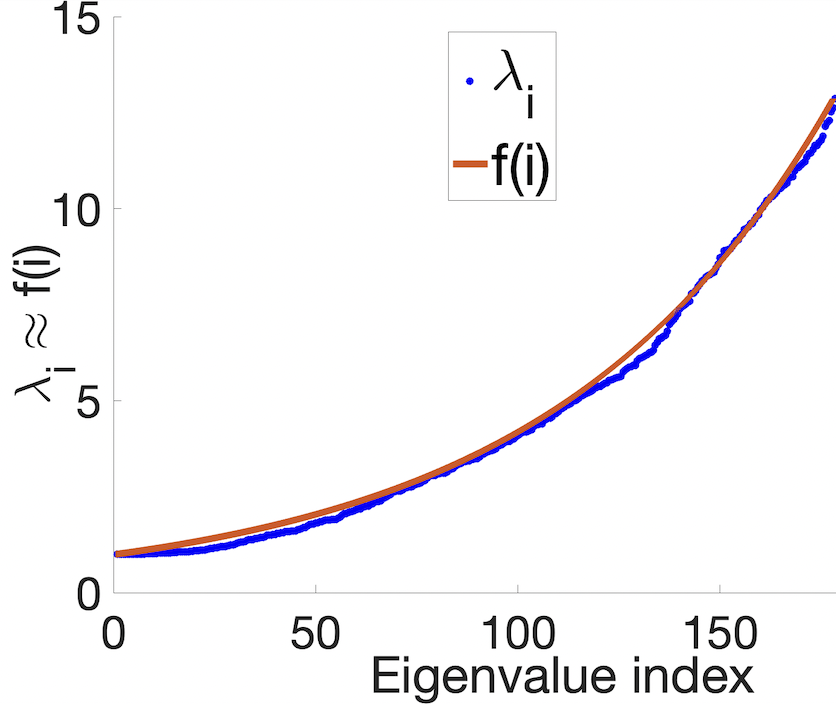}
\label{fig:pcd-on}}
\subfloat[{}]{
\includegraphics[width=0.235\textwidth]{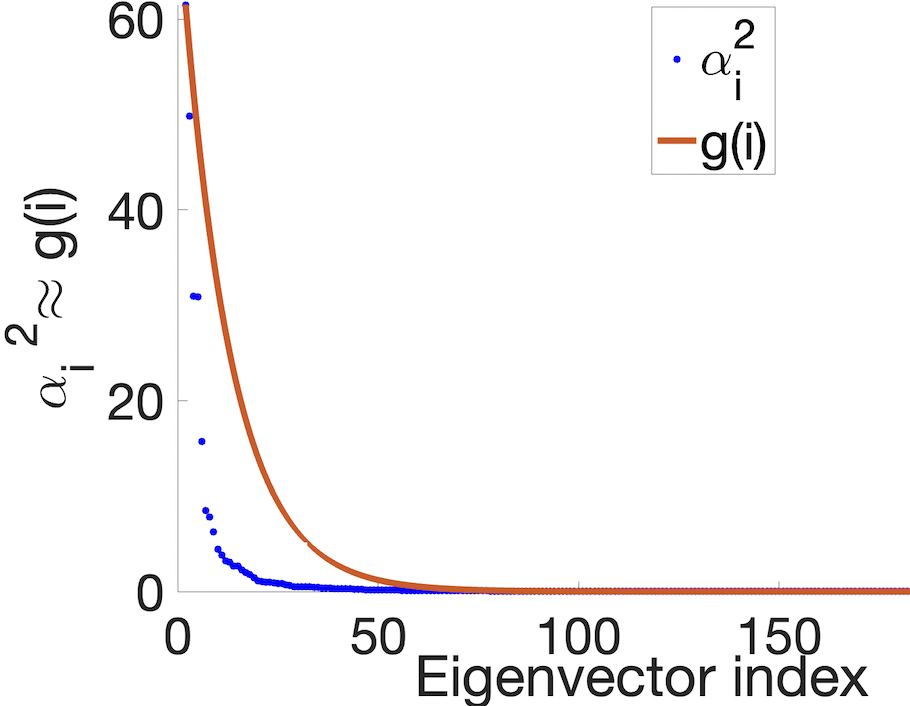}
\label{fig:pcd-off}}
\vspace{-0.05in}
\caption{(a) Modeling $\lambda_i$'s as an exponentially increasing function $f(i)$; (b) modeling $\alpha_i^2$ as an exponentially decreasing function $g(i)$.} 
\label{fig:approximations}
\end{figure}

Having estimated a noise variance $\sigma^2$, we now derive the optimal weight parameter $\mu$ for MAP formulation \eqref{eq:MAP}.
Following the derivation in \cite{chen17}, given $\sigma^2$, the mean square error (MSE) of the MAP estimate $\x^*$ from \eqref{eq:MAP} computed using ground truth signal $\x^o$ as function of $\mu$ is
\begin{align}
\text{MSE}(\mu) = \underbrace{\sum_{i=2}^N \psi_i^2 (\v_i^\top \x^o)^2}_{B(\mu)} + \underbrace{\sigma^2 \sum_{i=1}^N \phi_i^2}_{V(\mu)} 
\label{eq:MSE}
\end{align}
where $\psi_i = \frac{1}{1 + \frac{1}{\mu \lambda_i}}$ and $\phi_i = \frac{1}{1 + \mu \lambda_i}$. 
The first term $B(\mu)$ corresponds to the \textit{bias} of estimate $\x^*$, which is a differentiable, \textit{concave} and monotonically increasing function of $\mu>0$.
In contrast, the second term $V(\mu)$ corresponds to the \textit{variance} of $\x^*$, and is a differentiable, \textit{convex} monotonically decreasing function of $\mu>0$. 
When combined, MSE is a differentiable and provably \textit{pseudo-convex} function of $\mu>0$ \cite{mangasarian1975pseudo}, \ie, 
\begin{align}
\nabla \text{MSE}(\mu_1) \cdot (\mu_2 - \mu_1) \geq 0 \rightarrow \text{MSE}(\mu_2) \geq \text{MSE}(\mu_1),
\label{eq:covex_def}
\end{align}
$\forall \mu_1, \mu_2 >0$.
See Fig.\;\ref{fig:functions} for an example of bias $B(\mu)$, variance $V(\mu)$ and $\text{MSE}(\mu)$ for a specific graph signal $\x^o$ and a graph $\cG$, and Appendix~\ref{app:convex} for a proof of pseudo-convexity. 

In \cite{chen17}, the authors derived a corollary where $\text{MSE}(\mu)$ in \eqref{eq:MSE} is replaced by a convex upper bound $\text{MSE}^+(\mu)$ that is more easily computable.
The optimal $\mu$ is then computed by minimizing the convex function $\text{MSE}^+(\mu)$ using conventional optimization methods. 
However, this upper bound is too loose in practice to be useful.

Instead, we take an alternative approach: we approximate \eqref{eq:MSE} by modeling the distributions of eigenvalues $\lambda_i$'s of $\L$ and signal energies $\alpha_i^2 = (\v_i^\top \x^o)^2$ at graph frequencies $i$ as follows. 
We model $\lambda_i$'s as an exponentially increasing function $f(i)$, and model $\alpha_i^2$ as an exponentially decreasing function $g\left(i\right)$, namely
\begin{equation}
    \lambda_i\approx f(i)=q\exp\{\gamma i\}; \hspace{10pt} \alpha_i^2\approx g(i)=r\exp\{-\theta i\},
    \label{eq:approx_func}
\end{equation}
where $q, \gamma, r, \theta$ are parameters. 
See Fig.\;\ref{fig:approximations} for illustrations of both approximations.  
To compute those parameters, we first compute extreme eigen-pairs $(\lambda_i, \v_i)$ for $i \in \{2, N\}$ in linear time using LOBPCG~\cite{knyazev2001}. 
Hence we have following expressions from~(\ref{eq:approx_func}),
\begin{equation}
\begin{split}
    \lambda_2 &\approx q\exp\{2\gamma\}; \hspace{20pt} \lambda_N \approx q\exp\{N\gamma\},\\ \alpha_2^2 &\approx r\exp\{-2\theta\}; \hspace{15pt} \alpha_N^2 \approx r\exp\{-N\theta\}. 
\end{split}    
\end{equation}
By solving these equations, one can obtain the four parameters as
\begin{equation}
\begin{split}
    \gamma &= \frac{\ln{\frac{\lambda_N}{\lambda_2}}}{N-2}; \hspace{15pt} q=\lambda_2\exp\left\{-2\frac{\ln{\frac{\lambda_N}{\lambda_2}}}{N-2}\right\},\\
    \theta &= -\frac{\ln{\frac{\alpha_N^2}{\alpha_2^2}}}{N-2}; \hspace{10pt} r=\alpha_2^2\exp\left\{-2\frac{\ln{\frac{\alpha_N^2}{\alpha_2^2}}}{N-2}\right\}.
\end{split}
\end{equation}
One can thus approximate MSE in~(\ref{eq:MSE}) as
\begin{align}
\text{MSE}^a(\mu) = 
\sum_{i=2}^N \frac{g(i)}{\left(1+\frac{1}{\mu f(i)}\right)^2}
+ \sigma^2 \sum_{i=1}^N \frac{1}{\left(1 + \mu f(i)\right)^2} .
\label{eq:approx_MSE}
\end{align}

Since MSE in~(\ref{eq:MSE}) is a differentiable and pseudo-convex function for $\mu >0$, $\text{MSE}^a$ in~(\ref{eq:approx_MSE}) is also a differentiable and pseudo-convex function for $\mu>0$ with its gradient equals to
\begin{equation}
 \nabla \text{MSE}^a(\mu)=\sum_{i=2}^{N}\frac{2\mu g(i)f(i)^2-2f(i)\sigma^2}{(1+\mu f(i))^3}.   
\end{equation}
Finally, the optimal $\mu>0$ is computed by iteratively  minimizing the pseudo-convex function~$\text{MSE}^a(\mu)$ using a standard gradient-decent algorithm:
\begin{equation}
    \mu^{(k)}=\mu^{(k-1)}-t\nabla \text{MSE}^a(\mu^{(k-1)}),
\label{eq:GD}    
\end{equation}
where $t$ is the step size and $\mu^{(k)}$ is the value of $\mu$ at the $k$-th iteration. 
We compute \eqref{eq:GD} iteratively until convergence.

\section{Experimentation}
\label{sec:results}
\begin{figure}[t]
\centering
\hspace{-35pt}
\subfloat[]{
\includegraphics[width=0.26\textwidth]{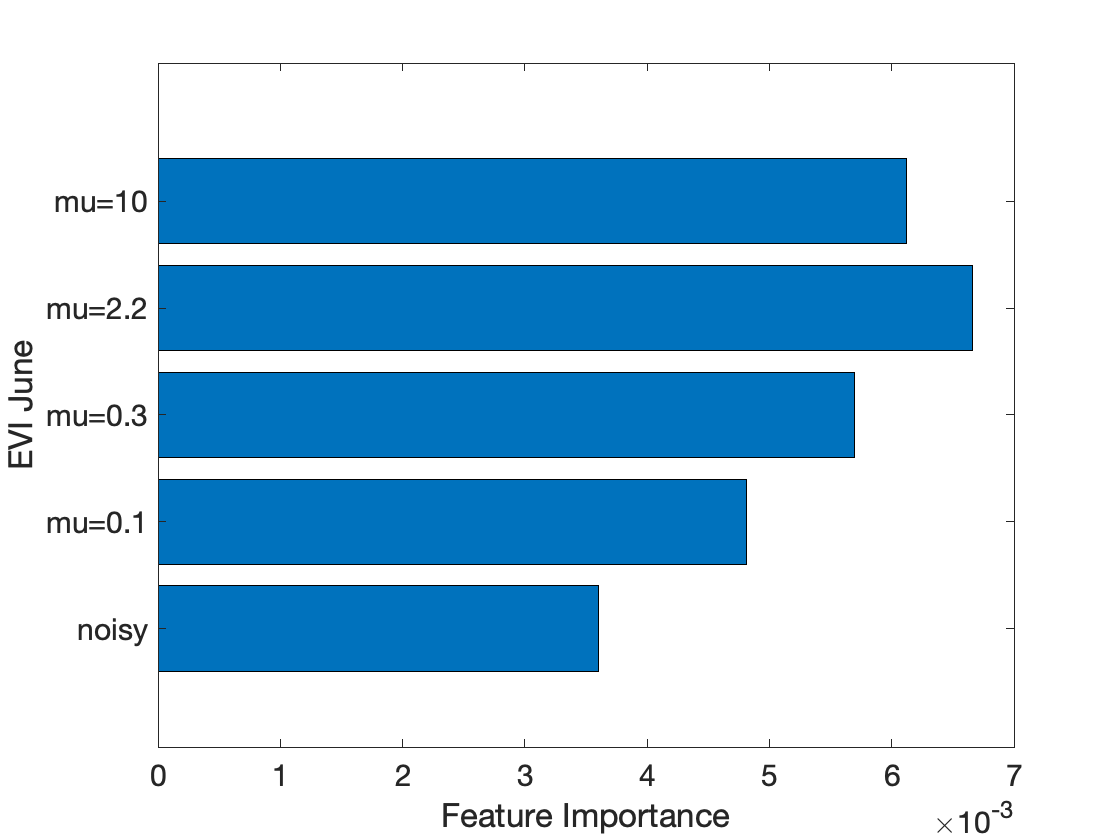}
\label{fig:pcd-on}}
\hspace{-0.25in}
\subfloat[{}]{
\includegraphics[width=0.26\textwidth]{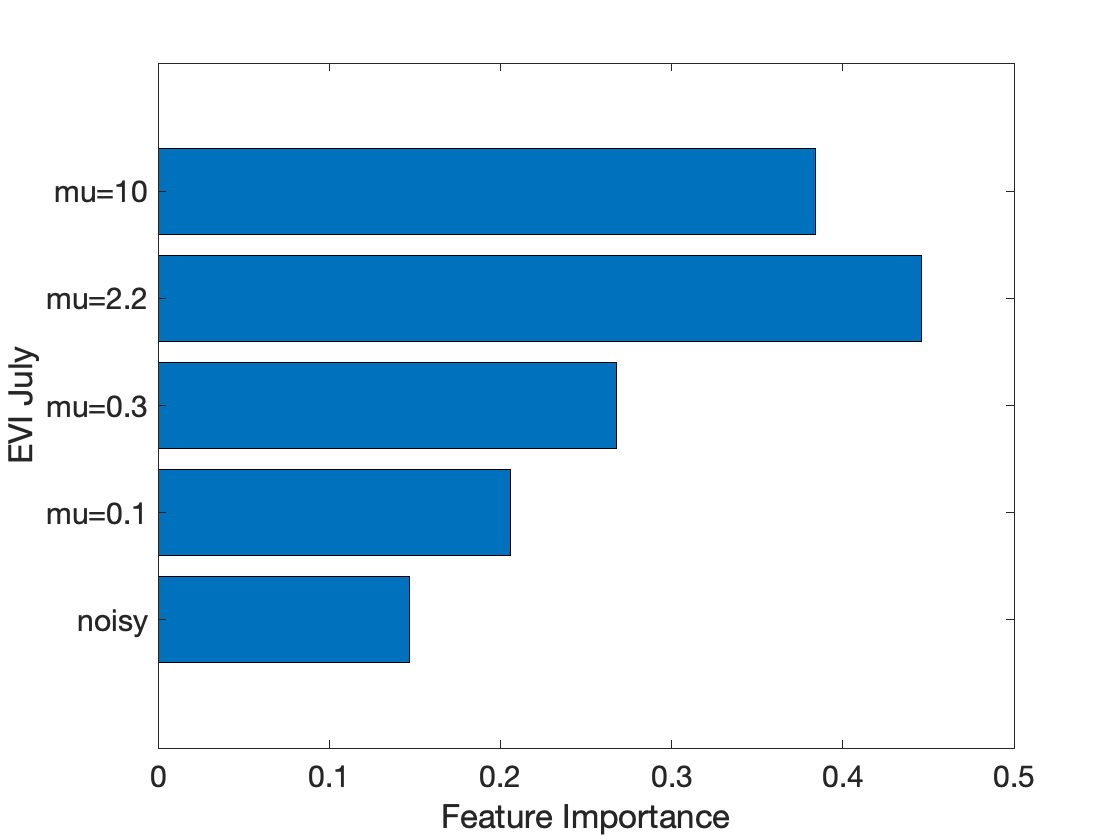}
\label{fig:pcd-off}}
\vspace{-0.05in}
\caption{ Feature importance for (a) \texttt{EVI\_June} and (b) \texttt{EVI\_July}.} 
\label{fig:featureimportance}
\vspace{-14pt}
\end{figure}

\begin{figure}[t]
\centering
\hspace{-30pt}
\subfloat[]{
\includegraphics[width=0.235\textwidth]{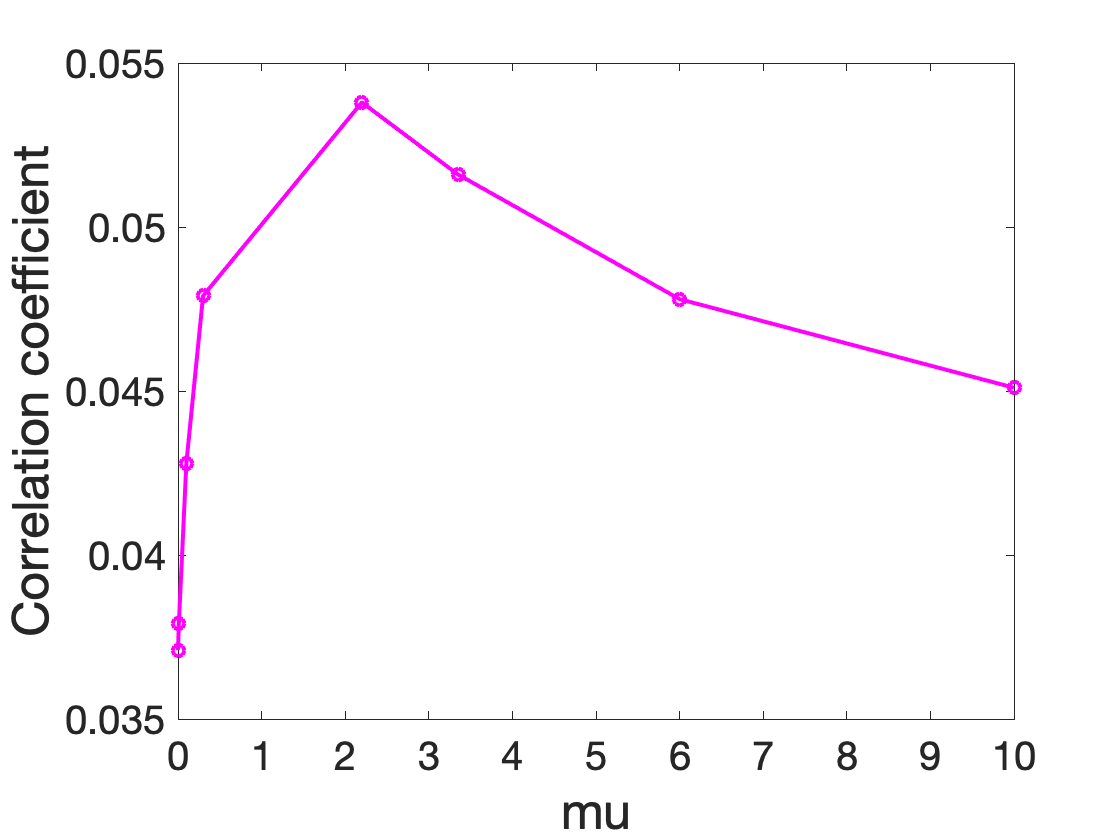}
\label{fig:pcd-on}}
\subfloat[{}]{
\includegraphics[width=0.235\textwidth]{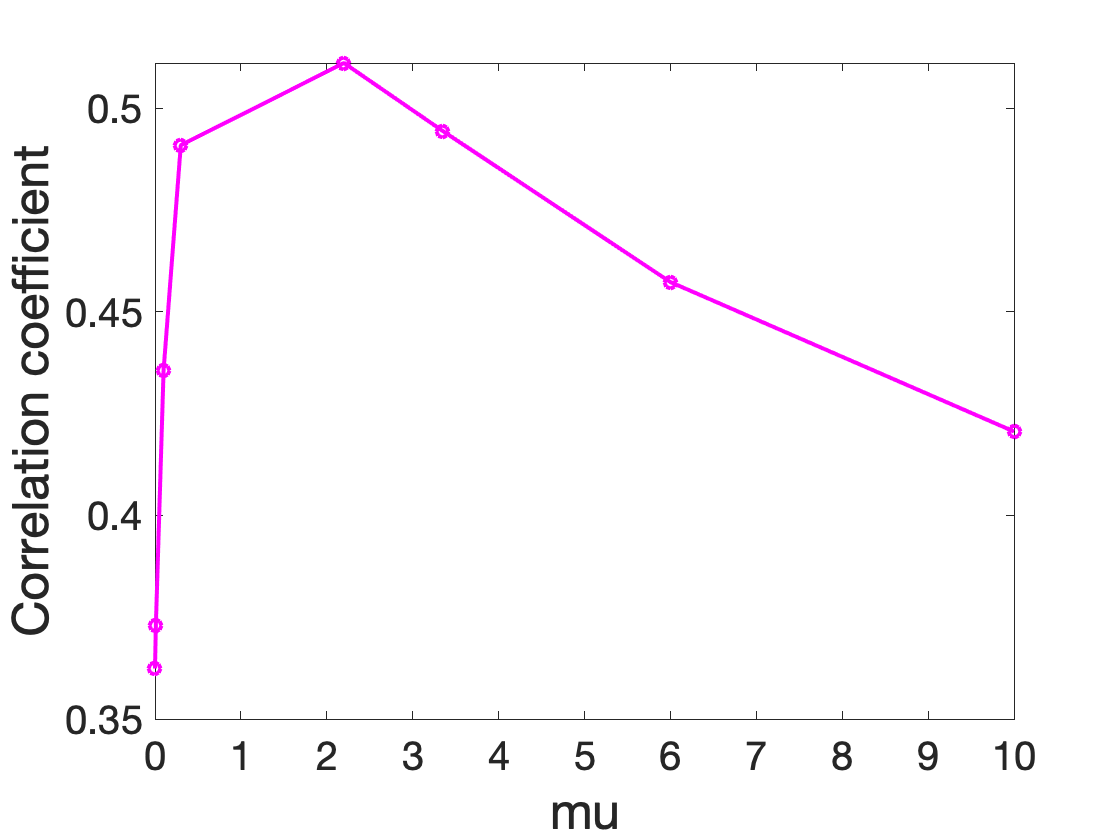}
\label{fig:pcd-off}}
\vspace{-0.05in}
\caption{ Correlation coefficient between denoised EVI feature ((a) \texttt{EVI\_June}; (b) \texttt{EVI\_July}) and the crop yield feature.} 
\label{fig:correlations}
\vspace{-10pt}
\end{figure}

\subsection{Experimental Setup}

To test the effectiveness of our proposed feature pre-denoising algorithm, we conducted the following experiment. 
 We used the corn yield data at the county level between year 2010 and 2019 provided by USDA and the National Agricultural Statistics Service\footnote{https://quickstats.nass.usda.gov/} to predict yields in 2020. 
We performed our experiments in $938$ counties in $10$ states (Iowa, Illinois, Indiana, Kentucky, Michigan, Minnesota, Missouri, Nebraska, Ohio, Wisconsin) in the corn belt. 
As discussed in Section\;\ref{sec:graph_learning}, we used five soil- and location-related features to compose feature $\f_i \in \mathbb{R}^5$ for each county $i$ and metric learning algorithm in \cite{wei_TSP2020} to compute metric matrix $\M$, in order to build a similarity graph $\cG$. 
For feature denoising, we targeted enhanced vegetation Index (EVI) for the months of June and July, \texttt{EVI\_June} and \texttt{EVI\_July}. 
In a nutshell, EVI quantifies vegetation greenness per area based on captured satellite images, and is an important feature for yield prediction. 
EVI is noisy for a variety of reasons: low-resolution satellite images, cloud occlusion, etc. 
We built a DL model for yield prediction based on XGBoost \cite{chen16} as the baseline, using which different versions of \texttt{EVI\_June} and \texttt{EVI\_July} were injected as input along with other relevant features. 


\subsection{Experimental Results}

First, we computed the optimum weight parameter $\mu$ for both \texttt{EVI\_June} and \texttt{EVI\_July}, which was $\mu = 2.2$. 
Table\;\ref{table1} shows the crop yield prediction performance using noisy features versus denoised features with different weight parameters, under three metrics in the yield prediction literature: root-mean-square error (RMSE), Mean Absolute Error (MAE) and R2 score (larger the better) \cite{pasquel22}. 
Results in Table\;\ref{table1} demonstrate that our optimal weight parameter (\ie, $2.2$) has the best results among other $\mu$ values. 
Specifically, our denoised features can reduce RMSE by $0.434\%$. 

In addition to the metrics in Table\;\ref{table1}, we measured the \textit{permutation feature importance}~\cite{altmann2010permutation} for both \texttt{EVI\_June} and \texttt{EVI\_July} before and after denoising. 
Fig.\;\ref{fig:featureimportance} shows that the importance of these features increases after denoising, demonstrating the positive effects of our unsupervised feature denoiser. 
Specifically, the result for the optimal $\mu = 2.2$ induced the most feature importance.

\begin{table}[h!] 
\caption{Performance Metrics with different weight parameters}
\label{table1}
\centering
\begin{footnotesize}
\begin{tabular}{|l|l|l|l|l|}
\hline
\multirow{1}{*}{Metric} &
Original &
$\mu = 0.001$ &
$\mu= 0.01$ &
$\mu = 2.2$  \\
\hline
RMSE (bu/ac) & 14.139 & 14.1966 &  14.2042 & \textbf{14.0776}  \\
\hline
MAE (bu/ac) & 11.225 & 11.2635 & 11.2674  & \textbf{10.9839} \\
\hline
R2 & 0.5894 & 0.5860 & 0.5856  & \textbf{0.5929} \\
\hline
\end{tabular}
\end{footnotesize}
\end{table}

Further, we calculated the correlation between the original / denoised feature \texttt{EVI\_June} and \texttt{EVI\_July} and actual crop yield. 
Fig.\;\ref{fig:correlations} shows that the denoised features with the optimal $\mu = 2.2$ has the largest correlation with the crop yield feature. 
In comparison, using previous method in \cite{chen17} to estimate $\mu = 3.35$ resulted in a weaker correlation. 

\begin{figure}[ht]
\centering
{\includegraphics[width=0.5\textwidth]{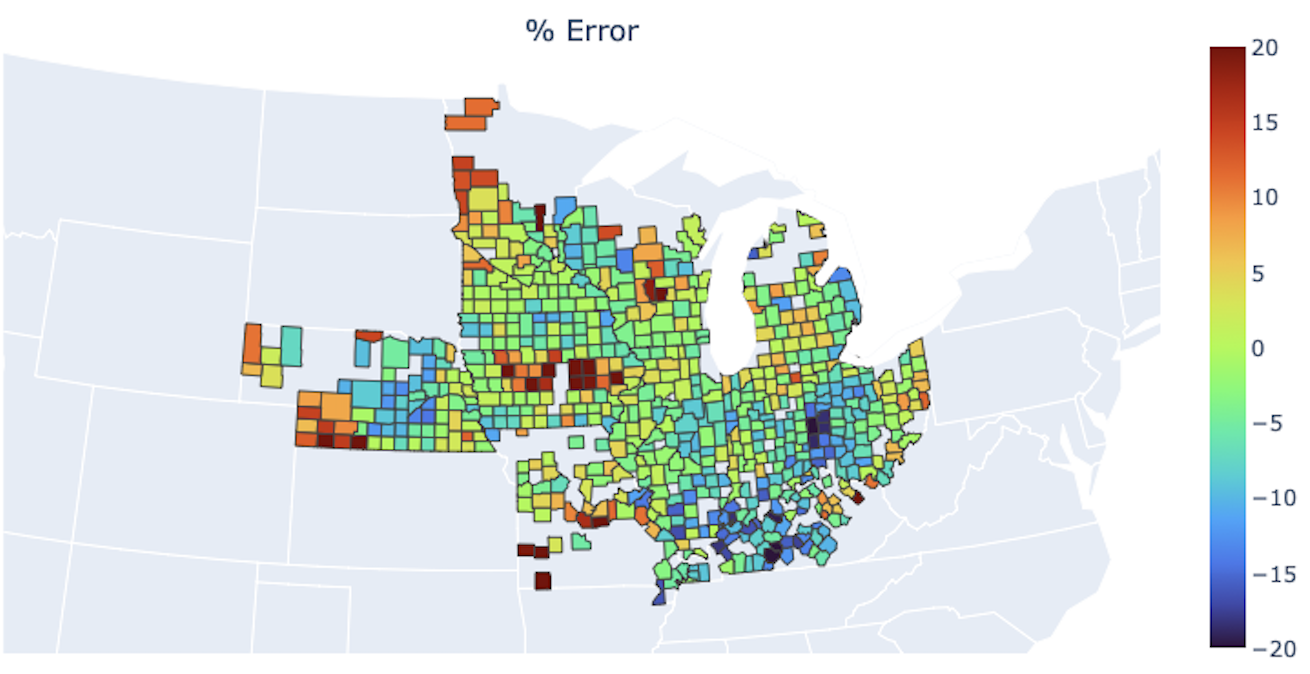}}
\vspace{-0.2in}
\caption{Yield prediction error in all the counties using denoised features}\label{fig:gl}
\label{fig:yield map}
\end{figure}

Lastly, to visualize the effect of our denoising algorithm, Fig.\;\ref{fig:yield map} shows the yield prediction error for different counties in the 10 states in the corn belt. 
We observe that with the exception of a set of counties in southern Iowa devastated by a rare strong wind event (called \textit{derecho}) in 2020, there were very few noticeably large yield prediction errors.

\section{Conclusion}
\label{sec:conclude}
Conventional crop yield prediction schemes exploit only temporal correlation to estimate future yields per county given input relevant features. 
In contrast, to exploit inherent spatial correlations among neighboring counties, we perform graph spectral filtering to pre-denoise input features for a deep learning model prior to network parameter training.  
Specifically, we formulate the feature denoising problem via a MAP formulation with the \textit{graph Laplacian regularizer} (GLR). 
We derive the weight parameter $\mu$ trading off the fidelity term against GLR in two steps.
We first estimate noise variance directly from noisy observations using a graph clique detection (GCD) procedure that discovers locally constant regions.
We then compute an optimal $\mu$ minimizing an MSE objective via bias-variance analysis.
Experiments show that using denoised features as input can improve a DL models' crop yield prediction.

\appendices

\section{Proof of pseudo-convexity for~(\ref{eq:MSE})}
\label{app:convex}

\begin{small}
We rewrite \eqref{eq:MSE}) as
\begin{equation}
    \text{MSE}(\mu)=\sum_{i=2}^{N}\frac{\mu^2\lambda_i^2\alpha_i^2 + \sigma^{2}}{(1+\mu\lambda_i)^2} + \sigma^2,
\end{equation}
where $\alpha_i^2 = (\v_i^\top \x^o)^2$. 
For simplicity, we only provide the proof for $N=i$. 
In this case, we can write,
\begin{equation}
 \nabla \text{MSE}(\mu)=\frac{2\mu \lambda_i^2\alpha_i^2-2\lambda_i\sigma^2}{(1+\mu \lambda_i)^3}.   
\end{equation}
Thus, the following expressions follow naturally:
\begin{equation}
\begin{split}
    \mu &\geq \frac{\sigma^2}{\lambda_i\alpha_i^2} > 0\rightarrow \nabla \text{MSE}(\mu)\geq 0; \\
    0 &<\mu < \frac{\sigma^2}{\lambda_i\alpha_i^2}\rightarrow \nabla \text{MSE}(\mu)< 0.
\end{split}
\label{eq:grad_sign}
\end{equation}
Further, according to~(\ref{eq:grad_sign}), for $\mu_1\geq\frac{\sigma^2}{\lambda_i\alpha_i^2} > 0$, 
\begin{equation}
  (\mu_2-\mu_1)\geq 0 \rightarrow (\text{MSE}(\mu_2)-\text{MSE}(\mu_1))\geq 0,
 \label{eq:grad_diff1} 
\end{equation}
and for $0<\mu_1<\frac{\sigma^2}{\lambda_i\alpha_i^2}$,
\begin{equation}
  (\mu_2-\mu_1)< 0 \rightarrow (\text{MSE}(\mu_2)-\text{MSE}(\mu_1))\geq 0.
  \label{eq:grad_diff2}
\end{equation}
Now, by combining \eqref{eq:grad_sign}, \eqref{eq:grad_diff1}, and \eqref{eq:grad_diff2}, one can write \eqref{eq:covex_def}, which concludes the proof.
\end{small}



\bibliographystyle{IEEEtran}
\bibliography{ref2}

\end{document}